\pdfoutput=1
\documentclass[runningheads]{llncs}
\usepackage[T1]{fontenc}

\usepackage{newtxtext}
\usepackage[varvw]{newtxmath}
\usepackage[table]{xcolor}
\usepackage{graphicx}
\usepackage{booktabs}
\usepackage{array}
\usepackage{adjustbox}
\usepackage[skins,raster]{tcolorbox}
\definecolor{qualBad}{HTML}{E97872}
\definecolor{qualGood}{HTML}{2E8B57}
\definecolor{qualText}{HTML}{143642}
\renewcommand*{\orcidID}[1]{\textsuperscript{\href{https://orcid.org/#1}{\includegraphics[height=2.5ex]{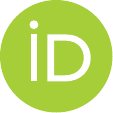}}}}

\newcommand{\module}[1]{\textsc{#1}}

\newcommand{\model}[1]{\textit{#1}}
\DeclareSymbolFont{cmsymbols}{OMS}{cmsy}{m}{n}
\DeclareMathSymbol{\cmrightarrow}{\mathrel}{cmsymbols}{"21}
\usepackage[hidelinks]{hyperref}

\begin{document}

\title{Prompt or Policy? Optimizing Physician Slate Presentations with LLM User Simulators}
\titlerunning{Optimizing Physician Slate Presentations with LLM User Simulators}

\author{Andrei Niculae\inst{1}\orcidID{0009-0002-5751-0514} \and
Adrian Cosma\inst{2}\orcidID{0000-0003-0307-2520} \and
Emilian Radoi\inst{1}\orcidID{0000-0002-1177-5288}}

\authorrunning{Niculae et al.}
\institute{National University of Science and Technology POLITEHNICA Bucharest, Romania
\and
IDSIA, Dalle Molle Institute for Artificial Intelligence, Lugano, Switzerland\\
Correspondence: \email{emilian.radoi@upb.ro}}

\maketitle
\begin{abstract}
Choosing the right physician is a key challenge for patients seeking medical care. Physician recommendation systems identify relevant doctors, but patients still find it difficult to compare their profiles to make a decision. We formulate \emph{slate presentation optimization}: given a patient's medical case and a fixed, ranked list of physicians, we generate short, case-specific summaries that make doctors easier to compare, while remaining grounded in the information in their profiles.
To optimize physician presentations without relying on real patient interactions, we develop an offline framework that uses persona-conditioned LLMs to simulate patient choices.
We generate patient personas from attributes extracted from each medical case and evaluate presentations for factual accuracy, relevance, naturalness and simulated patient choice.
We also use a second LLM judge, unseen during optimization, to assess simulator-specific overfitting.
We compare automatic prompt-level optimization with GEPA, parameter-level optimization with GRPO, and their combination across two model sizes.
GRPO substantially outperforms GEPA in satisfying the full set of constraints, and its gains in simulated conversion persist under the unseen judge.
GEPA alone brings smaller improvements, but using its optimized prompt to initialize GRPO speeds up training and achieves the highest simulated conversion.
These results show that prompt-level and parameter-level optimization play complementary roles in improving how physician recommendations are presented.

\keywords{Physician recommendation \and
Slate presentation optimization \and
Simulated patient choice \and
Automatic Prompt Optimization \and
RLAIF}
\end{abstract}

\section{Introduction and Related Work}
\label{sec:intro}
Choosing the right physician can be difficult, as patients often lack the medical knowledge needed to assess which doctor is best suited to their needs. Physician recommendation systems help address this problem by identifying relevant doctors based on patients' medical cases. Existing research has largely focused on retrieving and ranking physicians~\cite{lu-etal-2022-doctor,jiang2024online,zeng2026zero,liao2026patient}. However, even when relevant physicians are identified, patients must still interpret and compare information about their expertise, experience and patient feedback to make a decision \cite{shan2019influence,han2019exploring,gong2021factors,qin2022patient,ward2018trust,niculae-etal-2025-dr}.
In this work, we formulate \emph{slate presentation optimization}: given a patient's medical case and a fixed, ranked list of physicians, we generate concise, case-specific summaries that help patients understand and compare the doctors while remaining grounded in their profiles. The goal is to support patient choice, including the option not to consult any physician, without changing the ranking or making the decision for the patient.

Optimizing these presentations using real patient interactions is difficult: historical logs only capture responses to previously shown presentations~\cite{cosma2026improving}, while testing new alternatives requires live experiments.
User simulation \cite{rahdari2022ranked,bernard-2026-usersimcrs,zhang2024generative,balog2025theory} allows us to evaluate alternatives offline. LLM simulators can model richer patient behaviour, while persona conditioning captures differences in health literacy, urgency and affect \cite{kyung2026patientsim,mohtadi2026persona}. However, LLM simulators may be overly deterministic, insensitive to personas, or differ from real patients \cite{goli2024frontiers,califano2026large,kruff2026validating,meshi-etal-2026-convapparel,li2026llm}. We treat simulated conversion as an offline evaluation signal and test whether improvements persist under a second judge unseen during optimization. Synthetic feedback from LLM judges can optimize either the instructions (i.e., the prompt) or the model weights. We compare both approaches, as well as their composition, under the same constrained presentation objective. To optimize the prompt, \textit{Automatic Prompt Optimization} (APO) revises the instructions through examples, search, or textual feedback~\cite{pryzant-etal-2023-automatic,yang2024large,agrawal2026gepa,cosma2026automatic}; in our experiments, we use GEPA~\cite{agrawal2026gepa}. Such updates are lightweight, but their gains may depend on the base model and transfer poorly across model families~\cite{zhou2025evaluating,wang2025promptbridge}. To optimize the weights, Reinforcement Learning from AI Feedback (RLAIF) trains the model on the judges' scores. We use GRPO, which avoids a learned value model by comparing several outputs sampled for the same input~\cite{shao2024deepseekmath}. Compared with prompt optimization, this requires full training runs, and the model can learn to over-optimize an imperfect reward~\cite{gao2023scaling}.

Our framework evaluates each presentation for format compliance, faithfulness, relevance and safety, linguistic quality, topic coverage, and persona-conditioned simulated conversion. Across \model{gemma-3-4b} and \model{gemma-3-12b} \cite{gemmateam2025gemma3technicalreport}, GRPO satisfies the full constraint set substantially more reliably than GEPA alone. GEPA is nevertheless useful as a starting point for GRPO: the composition reaches high reward earlier and achieves the strongest simulated conversion, with gains persisting under the unseen judge.

\begin{figure}[t]
    \centering
    \includegraphics[width=0.85\linewidth]{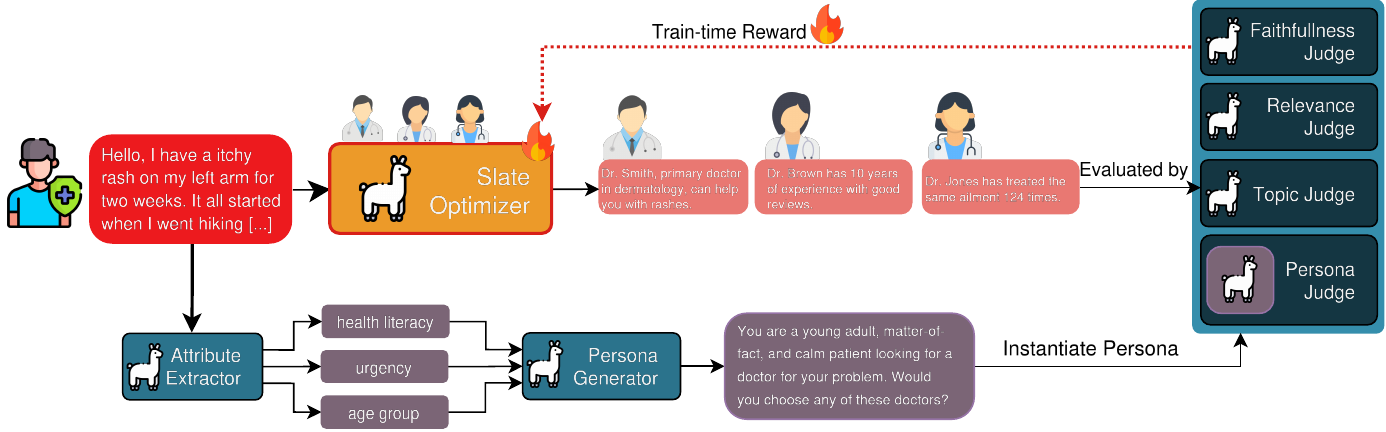}
    \caption{\textbf{Offline slate-presentation optimization.} \module{SlateOptimizer} summarizes a fixed physician slate, while case-derived attributes instantiate a simulated patient. Quality and persona-choice judges provide the reward for prompt-level (GEPA) or parameter-level (GRPO) optimization.}
    \label{fig:diagram}
\end{figure}

Our contributions are twofold:
\begin{enumerate}
    \item We formulate physician-slate presentation as a post-retrieval optimization problem with explicit factuality, safety and interface constraints, and introduce an offline protocol using persona-conditioned simulated choice and an unseen judge.
    \item We compare prompt-level, parameter-level and composed optimization across two model sizes, identifying their asymmetric roles: in-weight optimization enforces the constraints, while prompt optimization provides an effective initialization.
\end{enumerate}

\section{Method}
\label{sec:method}
An overview of our method is shown in Figure~\ref{fig:diagram}. Optimizing simulated conversion alone could reward persuasive but unsupported or unsafe text. Consequently, we introduce additional judges that act as regularizers of the slate presentation, ensuring the text is grounded in physician profiles. This suite of judges is used as reward for both GEPA and GRPO approaches.

\subsection{Suite of Judge LLMs for Slate Optimization}

\noindent\textbf{\module{SlateOptimizer}.}
Let $c$ denote the patient medical case and $D=(d_1,d_2,d_3)$ the fixed, ranked slate of physician profiles. Each profile contains structured and free-text evidence, including the physician's specialty and experience, education, response time, patient feedback, and previous consultations. \module{SlateOptimizer} maps $(c,D)$ to an ordered presentation $S=(s_1,s_2,s_3)$, where $s_i$ is a concise, case-specific summary of physician $d_i$. \module{SlateOptimizer} observes all three profiles jointly so that the cards can emphasize complementary evidence, but it may neither change the physicians nor reorder the slate.

\noindent\textbf{Constraint and quality judges.}
\module{HardChecks} assigns zero reward unless the output preserves physician order, keeps each card below 40 words, and avoids prohibited patterns such as superlatives, promises and urgency-inducing language. \module{FaithfulnessJudge} checks every claim against $c$ and the corresponding $d_i$; \module{RelevanceJudge} rates medical case relevance on a five-point scale and checks for medical advice; and \module{StyleJudge} rates naturalness on a five-point scale. A deterministic trigram-overlap score penalizes repetition. \module{TopicJudge} scores whether the patient's problem is contextually identified: 1 if relevant, 0 if mentioned irrelevantly and 0.5 if absent.

\noindent\textbf{Persona-conditioned choice judges.}
To model heterogeneous patient choices, we instantiate one simulated patient per medical case. Following PatientSim~\cite{kyung2026patientsim}, \module{AttributeExtractor} combines the reported mood with medical case-derived descriptors (patient role and age group, affect, urgency, health literacy and request type), and \module{PersonaGenerator} verbalizes them as a first-person prompt. These attributes remain hidden from \module{SlateOptimizer}, whose input is the original medical case.

\noindent\textbf{Reward and simulator transfer.}
\module{PersonaJudge} receives the persona and generated slate and rates willingness to consult any physician, with no consultation remaining valid. For outputs that pass \module{HardChecks}, this choice signal and the quality scores constitute the scalar reward. Since we are evaluating each training example under its associated persona, we are targeting performance across heterogeneous simulated patients. To test for simulator-specific overfitting, we also evaluate with a second choice judge never exposed during optimization (\model{Claude Haiku 5.5}).

\subsection{Experimental Setup}
We used patient-reported medical cases from medical forums and a physician database. Physician profiles and patient medical cases are written in Romanian. For each medical case, a retrieval model ranks physicians from this database, and we select the top three to form a slate.
We sample 10,000 slates, stratified by specialty, and split them into 8,000 training, 1,000 validation and 1,000 test slates. We used \model{gemma-3-4b} and \model{gemma-3-12b} \cite{gemmateam2025gemma3technicalreport} as the backbones of two \module{SlateOptimizer} variants. Each \module{SlateOptimizer} is a DSPy \cite{khattab2024dspy} program with a fixed seed instruction, which is also the unoptimized baseline. GRPO trains the weights behind this same prompt, so the two optimizers can be composed. We used \model{gemma-4-31b} \cite{gemmateam2026gemma4technicalreport} as the underlying LM for the judge modules and as a reflection model for GEPA. All experiments run on NVIDIA H200 GPUs. We run GEPA with a budget of 4,000 metric calls and reflection minibatches of 5 slates drawn from 300 train slates. The first 100 validation slates form the Pareto set on which candidates are selected. For GRPO, we fully fine-tune the two \module{SlateOptimizer} variants in prime-rl \cite{primerl2025}, using AdamW optimizer~\cite{kingma2017adammethodstochasticoptimization} with a constant learning rate of $10^{-6}$ (4B) and $5\cdot10^{-7}$ (12B). Each step samples 8 rollouts for each of 32 train slates (256 rollouts). We train for 125 gradient steps and evaluate the final checkpoint.

\noindent\textbf{Evaluation.} For each \module{SlateOptimizer} variant, we compare the unoptimized prompt, GEPA, GRPO and two compositions. These compositions are inspired by prior work showing that prompt and weight optimization are complementary: alternating the two outperforms either alone~\cite{soylu-etal-2024-fine}. GEPA $\cmrightarrow$ GRPO runs GRPO initialized from the optimized GEPA prompt, and GRPO $\cmrightarrow$ GEPA reruns GEPA on the GRPO-trained model. Every final program is scored once on the 1,000 test slates, and a slate without a readable output scores 0.

\section{Results}
\label{sec:experiments-results}
\newcommand{\keptnote}{$^\dagger$GEPA kept the GRPO prompt (none of its candidates beat it on the Pareto set), so GRPO $\cmrightarrow$ GEPA is the GRPO program.}

\begin{figure}[!hbt]
    \centering
    {\small\textbf{(a) Test performance}\par}
    \includegraphics[width=0.85\linewidth]{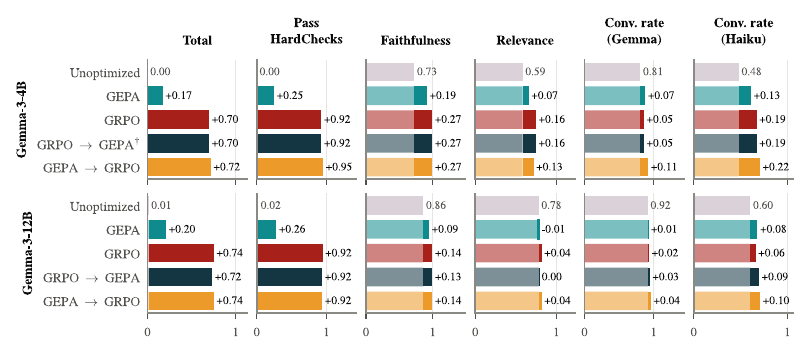}

    {\small\textbf{(b) GRPO training reward}\par}
    \includegraphics[width=0.95\linewidth]{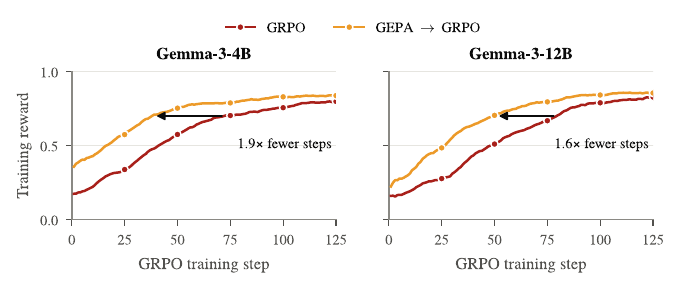}
    \caption{\textbf{Quantitative results.} \textbf{(a)} Test performance on 1,000 slates. The first bar is the unoptimized prompt (absolute value); subsequent bars show its difference. Pass \module{HardChecks}: fraction of slates that pass it with every claim supported. Conversion: fraction rated 4--5 by the training (\model{gemma}) or unseen (\model{Claude Haiku}) persona judge. \textbf{(b)} GRPO training reward from the original or GEPA-optimized prompt. \keptnote}
    \label{fig:quantitative-results}
\end{figure}

\noindent \textbf{Quantitative results.}
In Figure~\ref{fig:quantitative-results}(a) we report the test results of each final program. Before optimization, both \module{SlateOptimizer} variants already write slates on which the training persona judge would consult a physician. Almost none of these slates pass \module{HardChecks} with every claim supported, and most of the gains in total score come from satisfying these constraints: GRPO raises the pass rate from below 2\% to over 90\% for both variants, while the training judge's conversion rate changes only slightly. Prompt optimization improves individual criteria, such as faithfulness and relevance, but rarely satisfies all of them at once: GEPA reaches a pass rate of at most 28\%. GRPO learns the constraints directly from the reward. In-weight optimization also largely removes the advantage of the larger model. Before optimization, \model{gemma-3-12b} is better on every criterion; after GRPO, both variants perform similarly. In practice, a model three times smaller suffices, which lowers the cost of self-hosting, a relevant option for platforms that handle patient data.

In our experiments, we evaluate both composition directions, because we expected GRPO to mainly teach the hard checks, leaving room for a subsequent GEPA run to improve the quality and conversion judges. We find, however, that for 4B none of GEPA's candidates outperformed the GRPO prompt. The reverse order does help: starting GRPO from the GEPA prompt reaches the same training reward in 1.6--1.9$\times$ fewer steps (Figure~\ref{fig:quantitative-results}(b)). The reason is that a slate failing \module{HardChecks} receives a low reward, and GRPO computes advantages relative to the other rollouts for the same slate, so a group in which all rollouts fail provides no learning signal. A prompt that passes the checks more often yields more informative groups early in training.

We also observe that the conversion judge has a regularization effect for the quality judges during optimization. In preliminary runs where we used the conversion judge only for evaluation, the optimized cards satisfied the quality criteria but the conversion rate decreased. Having the conversion judge as part of the reward function, conversion remains high while the constraint and quality scores increase. The gains also transfer to the external judge: under \model{Claude Haiku 5.5}, which never provided a training signal, every optimized program raises the conversion rate, and GEPA $\cmrightarrow$ GRPO brings both variants to the same rate.

\noindent \textbf{Qualitative examples.}
In Table~\ref{tab:qualitative} we show two test slates for \model{gemma-3-12b}. The unoptimized cards describe the physician's profile generically, often with expertise unrelated to the medical case (diabetes for a feverish toddler) or with promotional wording (``natural beauty of your skin''). The optimized cards name the patient's problem and then cite the same two pieces of evidence: the share of positive patient votes and the typical reply time. These are among the factors that prior work identifies as influencing patients' choice of physician \cite{shan2019influence,han2019exploring,gong2021factors}.

\begin{table}
    \centering
    \caption{Qualitative \model{gemma-3-12b} examples before and after GEPA $\cmrightarrow$ GRPO. Medical cases are paraphrased; cards are translated from Romanian and anonymized. Persona: judge-only attributes. Judge: \model{Claude Haiku 5.5}'s unseen appeal score (1--5).}
    \label{tab:qualitative}
    \resizebox{0.9\textwidth}{!}{%
    \begin{minipage}{\textwidth}
    \scriptsize
    \tcbset{qualpart/.style={enhanced, arc=2.4pt, boxrule=0pt, left=4pt, right=4pt, top=3pt, bottom=3pt,
        boxsep=0pt, before skip=0pt, after skip=0pt, coltext=qualText},
        qualscore/.style n args={3}{bottom=12.5pt, overlay={
            \node[anchor=base west, inner sep=0pt, text=#1!80!black]
                at ([xshift=4pt, yshift=4.5pt]frame.south west) {\scriptsize\bfseries #3};
            \node[anchor=base east, inner sep=0pt, text=#1!80!black]
                at ([xshift=-4pt, yshift=4.5pt]frame.south east) {\scriptsize Judge: \textbf{#2/5}};}}}
    \newcommand{\qualexample}[6]{%
        \begin{tcolorbox}[enhanced, arc=3pt, boxrule=0.6pt, colframe=black!30, colback=white,
                left=0pt, right=0pt, top=0pt, bottom=0pt, boxsep=0pt, before skip=0pt, after skip=0pt]
            \begin{tcolorbox}[qualpart, sharp corners, rounded corners=north, colback=black!4, colframe=black!4, borderline south={0.6pt}{0pt}{black!30}]
                \raggedright\textbf{Patient:} #1\par\vspace{2pt}
                \textbf{Persona:} #2
            \end{tcolorbox}
            \begin{tcbraster}[raster columns=2, raster column skip=0pt, raster equal height=rows, raster before skip=0pt, raster after skip=0pt]
                \begin{tcolorbox}[qualpart, sharp corners, rounded corners=southwest, qualscore={qualBad}{#5}{Unoptimized},
                        colback=qualBad!14, colframe=qualBad!14, borderline east={0.3pt}{0pt}{black!30}]
                    \raggedright``#3''
                \end{tcolorbox}
                \begin{tcolorbox}[qualpart, sharp corners, rounded corners=southeast, qualscore={qualGood}{#6}{GEPA $\cmrightarrow$ GRPO},
                        colback=qualGood!16, colframe=qualGood!16, borderline west={0.3pt}{0pt}{black!30}]
                    \raggedright``#4''
                \end{tcolorbox}
            \end{tcbraster}
        \end{tcolorbox}\par\vspace{3pt}}
    \qualexample{Since last week my 15-month-old son has had a fever of 38.7--39\,\textdegree C that comes back after ibuprofen and very swollen gums, and since yesterday diarrhea; his head is very hot and he refuses food. Please help me with some advice and treatment.}
        {parent, child, worried, mood: sad, high urgency, low health literacy, treatment advice}
        {This pediatrician has expertise in diabetes and nutrition, offering consultations and glucose monitoring systems. They have helped many patients, with 12 positive votes, and have been on the platform for over a year.}
        {For your child's problems, you can consult this pediatric specialist. They get 100\% positive votes from patients and reply relatively quickly.}
        {1}{5}
    \qualexample{I am 35, and for about two years I have had a 3--4\,mm mole on my forearm, darker than my others. It does not bother me and has not changed, but I would like an opinion; I will attach a photo.}
        {self, adult, calm, low urgency, medium health literacy, what is it}
        {This dermatologist, also specialized in dermatoscopy and aesthetics, offers you an approach centered on the health and natural beauty of your skin, combining classic treatments with modern therapies for long-term care.}
        {This dermatologist, specialized in dermatoscopy, can evaluate your mole. They have 98\% positive votes and usually reply in under 4 hours.}
        {2}{5}
    \end{minipage}%
    }
\end{table}

\section{Conclusions}
\label{sec:conclusions}
We introduced slate presentation optimization as a post-retrieval task that aims to help patients compare recommended physicians through concise, case-specific summaries grounded in their profiles. We developed an offline framework that combines presentation quality assessments with persona-conditioned LLM simulations of patient choice to optimize these summaries without relying on real patient interactions.

Our experiments show that parameter-level optimization with GRPO is substantially more effective than prompt-level optimization with GEPA at meeting the requirements of this task. GRPO increased the proportion of slates passing all constraints from below 2\% to over 90\%, while GEPA alone reached at most 28\%. This improvement came from satisfying the constraints; the already high simulated conversion changed little. However, prompt optimization proved valuable as an initialization for GRPO. Starting from a GEPA-optimized prompt reduced the training steps needed to reach the same reward by a factor of 1.6--1.9 and achieved the highest simulated conversion. Applying GEPA after GRPO did not improve the overall score. We also found that the smaller 4B model performed comparably to the 12B model after GRPO, suggesting that smaller models may be sufficient for this task.

The gains in simulated conversion also persisted under a second LLM judge that was not used during optimization. This suggests that the improvements are not specific to the training judge. However, both evaluations rely on LLM simulators, which may share biases or differ from real patients. Moreover, higher simulated conversion does not necessarily imply better physician choices.

\bibliographystyle{splncs04}
\bibliography{refs}

\end{document}